\pdfoutput=1
\def\year{2021}\relax
\documentclass[letterpaper]{article} 
\usepackage{aaai21}  
\usepackage{times}  
\usepackage{helvet} 
\usepackage{courier}  
\usepackage[hyphens]{url}  
\usepackage{graphicx} 
\usepackage{natbib}  
\usepackage{caption} 
\usepackage{todonotes}
\usepackage{listings}
\usepackage[frozencache=true,cachedir=.]{minted}
\usepackage{xcolor}
\usepackage{subcaption}
\usepackage{multirow}
\usepackage{amsmath}
\usepackage{amsfonts}

\title{Griddly: A platform for AI research in games.}
\author{
   Chris Bamford,\textsuperscript{\rm 1} Shengyi Huang,\textsuperscript{\rm 2} Simon Lucas,\textsuperscript{\rm 1} \\
}
\affiliations{
    \textsuperscript{\rm 1}Queen Mary University\\
    \textsuperscript{\rm 2}Drexel University\\
}

\begin{document}

\maketitle

\begin{abstract}

In recent years, there have been immense breakthroughs in Game AI research, particularly with Reinforcement Learning (RL). Despite their success, the underlying games are usually implemented with their own  preset environments and game mechanics, thus making it difficult for researchers to prototype different game environments. However, testing the RL agents against a variety of game environments is critical for recent effort to study generalization in RL and avoid the problem of overfitting that may otherwise occur. In this paper, we present Griddly as a new platform for Game AI research that provides a unique combination of highly configurable games, different observer types and an efficient C++ core engine.  Additionally, we present a series of baseline experiments to study the effect of different observation configurations and generalization ability of RL agents.

\end{abstract}

\section{Background}

Many prominent successes in research into Artificial  Intelligence (AI) have emerged from creating agents that can achieve high scores in video games such as Atari 2600 games \cite{bellemare_2013,badia_2020}, custom toy game environments \cite{gym_minigrid,perezliebana_2018}, or using wrappers around popular video game such as Starcraft \cite{vinyals_2019} DOTA 2 \cite{Berner2019Dota2W} and NetHack Learning Environment \cite{kttler_2020}

Designing and implementing game environments to test the ability of different algorithms, such as Reinforcement Learning (RL), to reason, generalize and plan can be complex and time consuming. Even simple environments require implementing a number of common components such as rendering, game mechanics and optimization. A few solutions have been developed to abstract away the implementation details of environments and present researchers with a simplified interface to concentrate on building the specifics of environments to test their algorithms. For example the General Video Game Framework (GVGAI) \cite{perezliebana_2018} provides a platform in which games can be defined by a {\em Video Game Description Language} (VGDL). VGDL contains a set of pre-defined instructions which can be combined together to create the mechanics of many games. The layout of levels can also be defined using a simple character-based 2D ASCII map. GVGAI is commonly used for research into general game playing \cite{torrado_2018,balla_2020,ye_2020,justesen_2017}, and procedural content generation \cite{drageset_2019,dharna_2020,khalifa_2020,khalifa_2016}. 

In addition to the complexities involved in generating game mechanics and optimizing the game engine to allow fast prototyping during experimentation, the {\em representation} of game states used can also determine several factors during training such as how the agents generalize to unseen environment objects \cite{hill_2019} and new levels \cite{ye_2020, balla_2020}. The {\em representation} of the states can also have a large impact on speed and memory usage during learning. RL from the pixels in rendered game frames requires significantly larger memory usage when using neural networks compared to if a simple one-hot or multi-class map representation of the game state is used.

The final consideration for game environments is that of how the agent or agents interact with the environment itself. Actions spaces differ across types of game environments, for example if the environment is an RTS game, the agent may need to provide coordinates in order to target certain actions, and available actions may differ across various units in the game. In single-player games, the agent may only need to provide a single value to control movement or rotation. In multi-player games the environment needs an interface which provides the ability to control only certain units.

In this paper we introduce “Griddly”, which provides a highly configurable and optimized platform for building grid-world games for artificial intelligence research. Game environments in Griddly, like GVGAI use a domain specific language known as {\em Griddly Description YAML} (GDY) which allows an unprecedented level of configurability in all of the key areas described above. Not only can GDY be used to create single player puzzle games like those in GVGAI, MiniGrid~\cite{gym_minigrid}, DMLab2D~\cite{beattie_2020} and other toy problems, but it can be used to create multi-agent and RTS style games with partial observability and complex resource systems. GDY also provides multiple built-in observation representations such as sprite based isometric rendering, simple shape-based tiles and minimal state vectors. In addition to the wide array of configurability that Griddly provides, the underlying Griddly engine is heavily optimized in both computational speed and memory usage by taking advantage of hardware accelerated rendering techniques. 

To showcase Griddly, we provide a simple baseline of example experiments from 10 GVGAI-style games with various mechanics, configurations and observation representations. For each of the 10 games, we train a simple environment-size agnostic RL Agent on 5 different in-built levels, using three different observation configurations. In total this baseline results in 150 separate experiments. In addition to these 150, we provide 6 {\em generalization} experiments where we train an egocentric partial observability agent on three levels and evaluate its performance on two unseen levels.

Code for reproducing the experiments are made publicly available\footnote{https://github.com/Bam4d/griddly-paper}, so are the results and videos of each of these experiments\footnote{https://wandb.ai/griddly}. 
Full documentation, usage examples and tutorials for using Griddly are also provided\footnote{https://griddly.readthedocs.io/en/latest/}.

The structure of the rest of this paper is as follows:
Firstly we present a breakdown of the main configuration sections of Griddly’s description language (GDY), showing examples of creating simple game mechanics, how an environment can be set up for egocentric partial observability and how actions spaces can be defined. 
Secondly we show the results of the 150 baseline experiments that are trained per-level and the 6 generalization experiments.
Thirdly we compare of the speed of generated states and memory usage between a number of game environments and their Griddly equivalent using GDY on the same hardware.
Finally we discuss further features that we plan to add to Griddly and how Griddly can be used in future research.

\section{Griddly}
\label{sec:griddly}

\begin{figure}
    \centering
    \includegraphics[width=.7\linewidth]{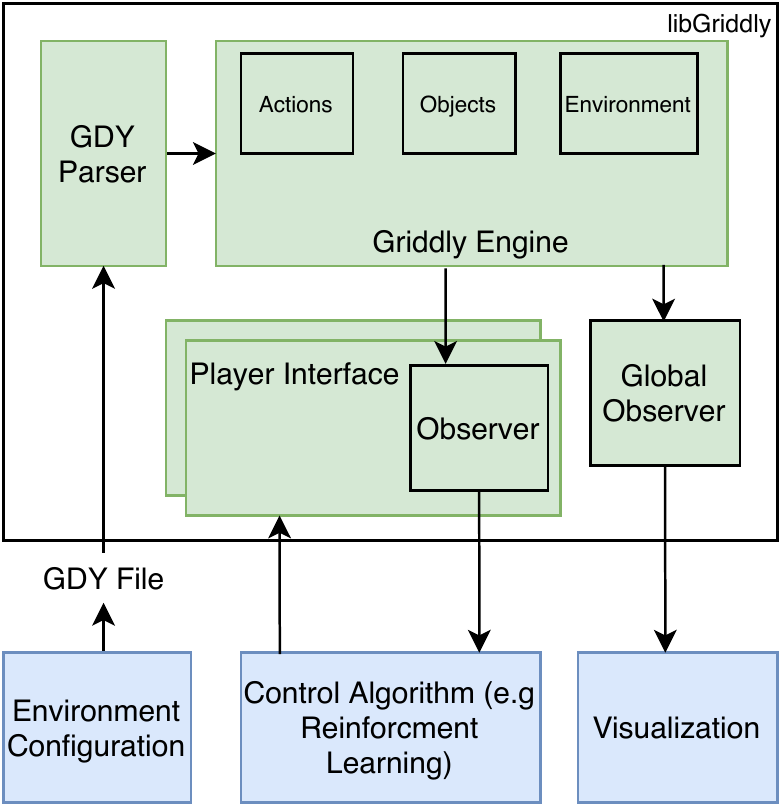}  
    \caption{A high level diagram of main components of the Griddly architecture showing the seperation of the main components. Multiple player interfaces with congifured observers can be attached to the Griddly engine to control any number of players in the environments, additionally a global observer can be used to monitor the environment as a whole or analyse the performance of any algorithms from a global perspective. The interface for configuring environments is also seperate from observation and agent control, meaning it can be used for generating algorithms for procedural content generation, or training with custom level designs.}
    \label{fig:griddly-engine}
\end{figure}

Griddly is an open-source project aimed to be a all-encompassing platform for grid-world based research. Griddly provides a highly optimized game state and rendering engine with a flexible high-level interface for configuring environments. Not only does Griddly offer simple interfaces for single, multi-player and RTS games, but also multiple methods of rendering, configurable partial observability and interfaces for procedural content generation.

Game mechanics are important building blocks of any game, they define the time and action dependent characteristics that turn a static image into an dynamic environment which can be used for understanding how fundamental logical understanding can help with much more complex tasks. As an example, the rules defined to allow an avatar to push a block one square at a time can be defined in a simple and deterministic way. This can be used to concentrate an AI agent learning to perform simple tasks such as collecting blocks in a specific area. In a physical world, this process would be significantly more complex. The block and the agent would be 3D, would have a high resolution view of the world, and there would be millions of variables to take into account, such as heat, friction, mass, forces of all the actuators of the agent etc. Using simplified game mechanics allows research to be undertaken on higher level concepts rather than have to take into account everything in the physical world. 

Griddly is designed to provide an simple YAML interface, Griddly Description Yaml (GDY) to allow many game mechanics to be defined in a simple way, without compromising on speed of rendering and mechanics. The next section will outline how the GDY schema can be used to create games.

\subsection{Griddly Description YAML (GDY)}

Griddly Description YAML (GDY) is a schema-oriented domain specific lanugage (DSL) which allows great flexibility in creating grid-world environments. Any DSL requires a certain amount of pre-existing knowledge, however with industry standard configuration languages there are many tools available such as syntax highlighting, schema validation and linting, which can reduce the barrier to entry when writing DSLs.
The use of YAML as a syntax allows schema validation  to be used for syntax highlighting, validation and auto-completion. Many IDEs such as Visual Studio Code, IntelliJ and PyCharm support YAML validation out-of-the box using JSON-schemas. This makes development of new games using Griddly simpler as the IDE will provide feedback on the game description's syntax and structure. The main GDY configurable features of Griddly games are the {\em Environment}, {\em Objects} and {\em Actions}. 

\subsubsection{Environment}

The environment Section of the YAML contains configuration options for several high level concepts for a Griddly game. The three most important of these are the {\em Player}, {\em Termination} and {\em Levels} options. The Player options define how the players will interact with the environment and which, if any, avatar object the player will control. Player partial observability can also be configured in the {\em Observer} subsection here using options such as: {\em RotateWithAvatar}, which causes the environment representation to rotate if the avatar rotates; {\em TrackAvatar}, which enables egocentric rendering (the agent is put at the center of the observation and {\em OffsetX} and {\em OffsetY} can be used to offset the agent from the center) and finally {\em Height} and {\em Width} which determine the height and width of the observable window. 

\begin{figure*}
    \centering
    \includegraphics[width=.5\textwidth]{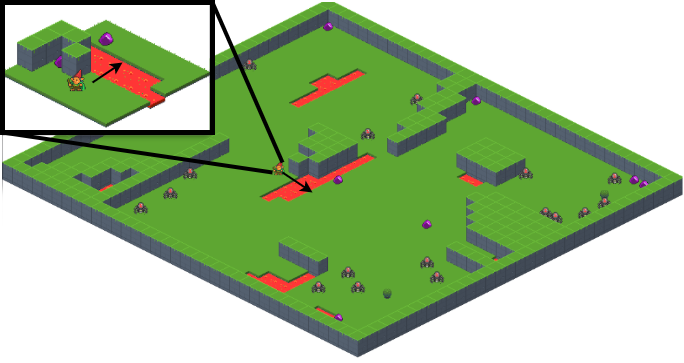}  
    \caption{An example of the rendering of the Griddly environment {\em spider nest} with a specific player's egocentric partially observable view that an agent will see during training (inset)}
    \label{fig:ego-part}
\end{figure*}

Termination conditions, such as determining if the episode is complete, or determining the winner in a multi-player game are also set in the environment section. Termination options can use any variable defined at a global level and also several special variables that allow calculations such as counts of specific objects in the environment. Levels are defined using strings of characters, the characters that are used are defined in the {\em Objects} configuration. An example of an environment configuration is shown below. An example of configured egocentric partial observability with an Isometric Renderer can be seen in figure \ref{fig:ego-part}

\begin{minted}
[
frame=lines,
framesep=2mm,
baselinestretch=1.2,
fontsize=\footnotesize,
linenos
]{yaml}
Environment:
  Name: sokoban
  TileSize: 24
  BackgroundTile: gvgai/newset/floor2.png
  Player:
    Observer:
      RotateWithAvatar: true
      TrackAvatar: true
      Height: 7
      Width: 7
      OffsetX: 0
      OffsetY: 2
    AvatarObject: avatar
  Termination:
    Win:
      - eq: [box:count, 0]
  Levels:
      - |
        wwwwwww
        w..hA.w
        w.whw.w
        w...b.w
        whbb.ww
        w..wwww
        wwwwwww
      - |
        wwwwwwwww
        ww.h....w
        ww...bA.w
        w....w..w
        wwwbw...w
        www...w.w
        wwwh....w
        wwwwwwwww
\end{minted}

\subsubsection{Objects}

Objects in environments are defined individually. The object definition allows individual objects to contain encapsulated variables, for example; hit points, resources and possessions such as keys. Rendering information is also defined per-object and passed to the rendering engine at run-time. An example of the object definition for a single avatar object is given below:

\begin{minted}
[
frame=lines,
framesep=2mm,
baselinestretch=1.2,
fontsize=\footnotesize,
linenos
]{yaml}
Objects:
- Name: avatar
   Z: 2
   MapCharacter: A
   Observers:
     Sprite2D:
       Image: images/gvgai/oryx/knight1.png
\end{minted}

\subsubsection{Actions}

Instead of having a fixed set of actions, GDY allows the user to define any number of actions and how they will interact with other objects. Actions have the ability to modify the encapsulated variables of objects, for example allowing hit points to be modified, or adding and removing objects. 

Actions in Griddly are defined in two parts, the \textbf{Input Mapping} and the \textbf{Behaviours}. The Input Mapping maps a set of distinct integers to a {\em Description}, {\em OrientationVector} and {\em VectorToDest}. The OrientationVector and VectorToDest are parameters that can then be used by the \textbf{Behaviours} to define how different objects react when the action is performed on them by another object. The object that is performing the action is referred to as the {\em source} object and object that is the target of the action is referred to as the {\em destination}. The following YAML snippet shows how the \textbf{InputMapping} of a particular action is represented in GDY.

\begin{minted}
[
frame=lines,
framesep=2mm,
baselinestretch=1.2,
fontsize=\footnotesize,
linenos
]{yaml}
Name: move
InputMapping:
  Inputs:
    1:
      Description: Rotate left
      OrientationVector: [-1, 0]
    2:
      Description: Move forwards
      OrientationVector: [0, -1]
      VectorToDest: [0, -1]
    3:
      Description: Rotate right
      OrientationVector: [1, 0]
  Relative: true
\end{minted}


In order for this Input Mapping to translate into a full action, the Behaviours of objects must be defined for this particular action. This is done by defining a list of commands that can happen to each object if they are the {\em source} or {\em destination} of the object. The source of an action is determined by the type of game that is being played. For example if the game is single player where the player controls a particular avatar, the source of the action is almost the avatar object (unless there are other actions which are automated such as random movement of enemies). In RTS-style games the source of the action is the object at the location selected by the player. An example of how Behaviours are defined in GDY is shown in below:

\begin{minted}
[
frame=lines,
framesep=2mm,
baselinestretch=1.2,
fontsize=\footnotesize,
linenos
]{yaml}
Behaviours:
  - Src:
      Object: agent
      Commands:
        - rot: _dir
    Dst:
      Object: agent

  - Src:
      Object: agent
      Commands:
        - mov: _dest
    Dst:
      Object: _empty
\end{minted}

When an action is performed, destination object is located by adding the value of VectorToDest to the source location. If the Input Mapping is defined with {\em Relative:true}, the VectorToDest and OrientationVector are calculated relative to the current orientation of the object.

\subsection{Observers}

\begin{figure}
    \begin{subfigure}{.3\linewidth}
        \centering
        \includegraphics[width=.9\textwidth]{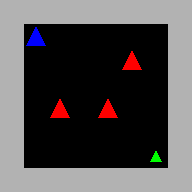}  
    \end{subfigure}%
    \begin{subfigure}{.3\linewidth}
    \centering
        \includegraphics[width=.9\textwidth]{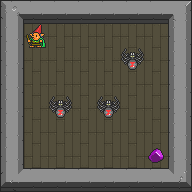}  
    \end{subfigure}%
    \begin{subfigure}{.3\linewidth}
        \centering
        \includegraphics[width=.9\textwidth]{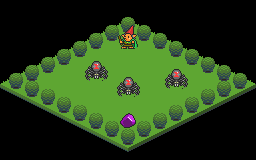}  
    \end{subfigure}                 
    \caption{Visualization of the three observers, {\em Block}, {\em Sprite} and {\em Isometric} configured on the "Spiders" environment. The 4th {\em Vector} observer does not have a natural visual representation.}
    \label{fig:observers}
\end{figure}

As Griddly supports multi-player, single-player, and RTS style games, the format of the data in which the players (whether they are algorithms or humans), can be specified individually. There are four available observer types: {\em Isomentric} {\em sprite}, {\em block} and {\em vector}. {\em Isometric}, {\em sprite}, {\em block} use hardware accelerated rendering through the Vulkan API \cite{vulkan}, an example of this is shown in Figure \ref{fig:observers}. {\em vector} rendering provides a lightweight class label representation of the grid. All observers support configurable partial observability and avatar tracking. Additionally multiple renderers can be used at the same time, meaning the game world can be observed separately from the agents. This allows rich demonstrations to be produced, even if the algorithms use the vector observer.

\subsection{Potential Applications}

GDY provides a flexible method of creating everything in a grid-world environment from game mechanics, to different ways of rending the state. Griddly comes pre-loaded with 21 initial game examples, many of which are in this paper. We wish to highlight a number of research areas where using Griddly would be beneficial.

\subsubsection{Environment Mechanics}

As games mechanics can be defined by combinations of simple instructions, many different game mechanics can be created. As an example, games that require random movement, ranged actions, directed movement (such as gravity) can be generated. Also resource systems can be easily defined by declaring variables that are local to objects or players, this means the possibility of complex games that require agents to collect particular items for specific tasks can be easily achieved. 

\subsubsection{Procedural Content Generation}

Due to the fact that the GDY language is YAML based, it is supported by most modern languages and there are many tools and libraries available to parse, edit and verify YAML files. The generation of game mechanics, levels and image assets can be performed by different projects regardless of language choices. The generated YAML can then be loaded by Griddly agnostically. 

\section{Baselines}

To provide future experiments with a RL baseline, we ran two different types of experiment. 

Firstly we chose 10 environments that have been ported from the GVGAI environment that are particularly difficult for RL algorithms to solve. Each environment contains 5 levels of different sizes and varying difficulty. We trained each level from each environment separately using three different observers {\em Vector}, {\em Block} and {\em Sprite}. Additionally two of the environments are configured with egocentric partial observability. This baseline suite consists of 150 experiments in total.  Each level is trained for 1.28M frames, and the average score for the 100 final episodes during training is measured. Table \ref{tab:single-env-results} shows the results of these experiments.

The second baseline uses a subset of the previous experiments, but all configured with egocentric partial observability. These experiments have an observation space of $5 \times 5$ tiles, with the agent centered at the bottom of this grid. This allows the agent to see 4 squares ahead of it in the direction of travel, and 2 squares either side.   Additionally, instead of training each level separately, 3 of the levels are used as a training set and then the 2 remaining levels are used to evaluate generalization. These experiments are all performed using only the {\em Vector} observer to produce game states.

The architecture of the neural network for both sets of experiments is the same. Both networks are trained using Proximal Policy Optimization \cite{schulman_2017} and Random Network Distillation \cite{burda_2018} (RND). The agent network, RND target network and RND prediction network share the same architecture per experiment. The only difference in networks occurs in the first few layers as the observation space differs between different observers used. Experiments using {\em Sprite} and {\em Block} observers begin with a convolutional layer with kernel size and stride the same as the tile size. This effectively embeds each tile into tensor $\mathbb{R}^{32 \times W \times H}$ where $C$ is the number of channels, and $H$ and $W$ are the height and width of the game level in terms of tiles. Alternatively experiments using {\em Vector} observers embed the vector representations using a $1 \times 1$ kernel with 32 channels into the same resulting shape $\mathbb{R}^{32 \times W \times H}$.
The embedded representation is then fed through the following layers: A convolutional layer with 32 input channels, 64 output channels, kernel size 3 and padding 1. Then a convolutional layer with 64 input and output channels, with kernel size 3 and padding 1. A global average pooling (GAP) layer is then used to reduce the network size to a linear layer of size 2048, this is then reduced by a further linear layer to a vector size 512. Single linear layers are used after this point for the various heads such as the actor and critic. 

The global average pooling layer allows different $H$ and $W$ for different level sizes to output the same shape vector (in our case 2048) \cite{balla_2020}.

The action space of the fully observable environments consists of 5 actions {\em do nothing (no-op)}, {\em move up}, {\em move left}, {\em move down}, and {\em move right}. In contrast, the partially observable setting the agent has access to 4 actions, {\em do nothing (no-op)}, {\em turn left}, {\em turn right} and {\em move forward}.  

\section{Results}

\subsection{Per-Environment}

\begin{table}[]

\resizebox{\linewidth}{!}{%
\begin{tabular}{c|c|c|ccc}
                         & \multicolumn{1}{c|}{\multirow{2}{*}{Level}} & \multicolumn{1}{c|}{Max}& \multicolumn{3}{c}{Observer} \\
Game                     &                                             & Reward & Vector   & Block   & Sprite  \\ \hline
\multirow{5}{*}{Clusters} & 0                                          & 8      & -0.06           $\pm$ 0.24 &      0.00            $\pm$ 0.00 &       2.00             $\pm$ 0.00                  \\
                         & 1                                           & 8      & -0.02           $\pm$ 0.14 &      3.00            $\pm$ 0.00 &       0.00             $\pm$ 0.00                  \\
                         & 2                                           & 6      &  0.00           $\pm$ 0.00 &      0.00            $\pm$ 0.00 &       0.00             $\pm$ 0.00                  \\
                         & 3                                           & 9      & -0.04           $\pm$ 0.20 &      1.00            $\pm$ 0.00 &       1.00             $\pm$ 0.00                  \\
                         & 4                                           & 6      &  0.00           $\pm$ 0.00 &      0.00            $\pm$ 0.00 &       0.00             $\pm$ 0.00                  \\ \hline
\multirow{5}{*}{Cook Me Pasta} & 0                                     & 25     &  0.08           $\pm$ 0.57 &      8.00            $\pm$ 0.00 &       0.00             $\pm$ 0.00                  \\
                         & 1                                           & 25     &  1.47           $\pm$ 1.93 &      8.00            $\pm$ 0.00 &       0.00             $\pm$ 0.00                  \\
                         & 2                                           & 25     & -0.04           $\pm$ 0.20 &      4.00            $\pm$ 0.00 &       0.00             $\pm$ 0.00                  \\ 
                         & 3                                           & 25     &  0.00           $\pm$ 0.00 &      0.00            $\pm$ 0.00 &       0.00             $\pm$ 0.00                  \\
                         & 4                                           & 25     &  0.00           $\pm$ 0.00 &      0.00            $\pm$ 0.00 &       0.00             $\pm$ 0.00                  \\ \hline
\multirow{5}{*}{Bait} & 0                                              & 5      &  5.00           $\pm$ 0.00 &      5.00            $\pm$ 0.00 &       5.00             $\pm$ 0.00                  \\
                         & 1                                           & 7      &  0.00           $\pm$ 0.00 &      1.00            $\pm$ 0.00 &       0.00             $\pm$ 0.00                  \\
                         & 2                                           & 12     &  1.00           $\pm$ 0.00 &      1.00            $\pm$ 0.00 &       1.00             $\pm$ 0.00                  \\ 
                         & 3                                           & 42     &  7.04           $\pm$ 0.28 &     23.00            $\pm$ 0.00 &      25.90             $\pm$ 0.36                  \\ 
                         & 4                                           & 12     &  7.00           $\pm$ 0.00 &      7.00            $\pm$ 0.00 &       7.00             $\pm$ 0.00                  \\ \hline
\multirow{5}{*}{Bait (Keys)} & 0                                       & 5      &  \textbf{5.00}  $\pm$ 0.00 &      \textbf{5.00}   $\pm$ 0.00 &       3.98             $\pm$ 2.02                  \\
                         & 1                                           & 7      &  0.00           $\pm$ 0.00 &      0.00            $\pm$ 0.00 &       0.00             $\pm$ 0.00                  \\
                         & 2                                           & 12     &  1.00           $\pm$ 0.00 &      1.00            $\pm$ 0.00 &       2.00             $\pm$ 0.00                  \\ 
                         & 3                                           & 42     &  7.00           $\pm$ 0.00 &      5.78            $\pm$ 0.65 &      26.90             $\pm$ 0.30                  \\
                         & 4                                           & 12     &  7.00           $\pm$ 0.00 &      6.69            $\pm$ 0.91 &       7.00             $\pm$ 0.00                  \\ \hline
\multirow{5}{*}{Sokoban} & 1                                           & 4      &  0.00           $\pm$ 0.00 &      0.00            $\pm$ 0.00 &       0.00             $\pm$ 0.00                  \\ 
                         & 2                                           & 3      &  0.00           $\pm$ 0.00 &      \textbf{3.00}   $\pm$ 0.00 &       2.00             $\pm$ 0.00                  \\
                         & 3                                           & 4      &  2.00           $\pm$ 0.00 &      \textbf{4.00}   $\pm$ 0.00 &       2.00             $\pm$ 0.00                  \\ 
                         & 4                                           & 3      &  2.00           $\pm$ 0.00 &      2.00            $\pm$ 0.00 &       1.00             $\pm$ 0.00                  \\
                         & 5                                           & 2      &  0.00           $\pm$ 0.00 &      \textbf{2.00}   $\pm$ 0.00 &       0.00             $\pm$ 0.00                  \\ \hline
\multirow{5}{*}{Sokoban 2} & 0                                         & 4      &  2.00           $\pm$ 0.00 &      2.00            $\pm$ 0.00 &       2.00             $\pm$ 0.00                  \\
                         & 1                                           & 3      &  0.00           $\pm$ 0.00 &      0.00            $\pm$ 0.00 &       0.00             $\pm$ 0.00                  \\ 
                         & 2                                           & 6      &  5.00           $\pm$ 0.00 &      4.92            $\pm$ 0.27 &       4.96             $\pm$ 0.20                  \\ 
                         & 3                                           & 2      &  1.00           $\pm$ 0.00 &      1.00            $\pm$ 0.00 &       1.00             $\pm$ 0.00                  \\
                         & 4                                           & 2      &  0.00           $\pm$ 0.00 &      0.00            $\pm$ 0.00 &       0.00             $\pm$ 0.00                  \\ \hline
\multirow{5}{*}{Zen Puzzle} & 0                                        & 34     & 27.24           $\pm$ 3.11 &     33.00            $\pm$ 0.00 &      \textbf{34.00}    $\pm$ 0.00                  \\
                         & 1                                           & 34     & 30.53           $\pm$ 1.74 &     \textbf{34.00}   $\pm$ 0.00 &      33.00             $\pm$ 0.00                  \\
                         & 2                                           & 33     & 26.14           $\pm$ 2.27 &     30.31            $\pm$ 1.68 &      28.98             $\pm$ 0.14                  \\ 
                         & 3                                           & 23     & 19.94           $\pm$ 0.24 &     18.00            $\pm$ 0.00 &      20.00             $\pm$ 0.00                  \\ 
                         & 4                                           & 27     & 26.90           $\pm$ 0.30 &     25.98            $\pm$ 0.14 &      25.00             $\pm$ 0.00                  \\ \hline
\multirow{5}{*}{Labyrinth} & 0                                         & 1      &  0.00           $\pm$ 0.00 &      0.00            $\pm$ 0.00 &       0.00             $\pm$ 0.00                  \\
                         & 1                                           & 1      & -0.04           $\pm$ 0.20 &      0.00            $\pm$ 0.00 &       0.00             $\pm$ 0.00                  \\
                         & 2                                           & 1      & -0.02           $\pm$ 0.14 &      0.00            $\pm$ 0.00 &       0.00             $\pm$ 0.00                  \\
                         & 3                                           & 1      &  0.00           $\pm$ 0.00 &      0.00            $\pm$ 0.00 &       0.00             $\pm$ 0.00                  \\ 
                         & 4                                           & 1      &  0.00           $\pm$ 0.00 &      0.00            $\pm$ 0.00 &       0.00             $\pm$ 0.00                  \\ \hline
\multirow{5}{*}{Labyrinth \textsuperscript{po}} & 0                    & 1      & \textbf{1.00}   $\pm$ 0.00 &      0.00            $\pm$ 0.00 &       0.00             $\pm$ 0.00                  \\
                         & 1                                           & 1      &  0.00           $\pm$ 0.00 &      0.00            $\pm$ 0.00 &       0.00             $\pm$ 0.00                  \\
                         & 2                                           & 1      & \textbf{1.00}   $\pm$ 0.00 &      0.00            $\pm$ 0.00 &       0.00             $\pm$ 0.00                  \\
                         & 3                                           & 1      &  0.00           $\pm$ 0.00 &      0.00            $\pm$ 0.00 &       \textbf{1.00}    $\pm$ 0.00                  \\
                         & 4                                           & 1      &  0.00           $\pm$ 0.00 &      0.00            $\pm$ 0.00 &       0.00             $\pm$ 0.00                  \\ \hline
\multirow{5}{*}{Zen Puzzle \textsuperscript{po}} & 0                   & 34     & \textbf{34.00}  $\pm$ 0.00 &     31.02            $\pm$ 1.67 &      30.00             $\pm$ 0.00                  \\
                         & 1                                           & 34     & 27.00           $\pm$ 0.00 &     32.00            $\pm$ 0.00 &      31.88             $\pm$ 0.48                  \\
                         & 2                                           & 33     & 29.00           $\pm$ 0.00 &     31.00            $\pm$ 0.00 &      28.00             $\pm$ 0.00                  \\
                         & 3                                           & 23     & 19.84           $\pm$ 0.37 &     19.90            $\pm$ 0.30 &      19.63             $\pm$ 0.48                  \\
                         & 4                                           & 27     & 22.00           $\pm$ 0.00 &     22.00            $\pm$ 0.00 &      22.00             $\pm$ 0.00                  \\ \hline
\end{tabular}}
\caption{This table shows the results of training 5 levels of 10 GVGAI environments that have been ported to the Griddly platform using GDY. Each level from each environment is trained for 1.28 million steps and the average episode reward of the final 100 epsodes is reported for each observer used. (Vector, Block and Sprite). In most cases the results are consistent across all the representations, but due to the unstable nature of RL based on random starting seeds, some errors in consistency are expected. The two environments marked with \textsuperscript{po} are configured with egocentric partial observability}
\label{tab:single-env-results}
\end{table}

As shown in table \ref{tab:single-env-results} there are levels of certain environments that fail to gain any score. These environments are sometimes difficult to solve even for humans as they require precise planning and making single mistakes result in states that cannot be reversed. For example in the "Zenpuzzle" the agent gains a reward every time it moves over tiles of a certain colour, but once it has moved over these blocks, it can not move onto them again. This can cause the agent to get "stuck" surrounded by tiles that it has already passed. The agent can also easily stop itself from being able to reach certain tiles by blocking the path to them. Although the agent scores highly in the "Zenpuzzle" environment, it rarely covers "all" the tiles to reach the maximum score possible. In the game "Clusters" the agent must push coloured blocks into groups and scores a point each time a block is "grouped". There are other obstacles in the environment that the agent must avoid pushing the blocks into. In the "Clusters" experiments the agent rarely learns to make a single group.

We test two partially observable versions of the games "Labyrinth" and "Zen Puzzle" to see if more general approaches, such as wall following strategies can be learned. As expected in "Labyrinth", a simple maze game with a reward at the destination, the agent performs better and can solve some of the mazes. We observe that in some of the games where a "wall following" approach can solve the maze, the agent learns this strategy. In other levels, the agent does not learn a strategy and fails to find the goal. It is also interesting to note that the full observable maze levels could not be learned by this method.

In the "Sokoban" environment, some levels are consistently solved. However, on other levels, the agent consistently fails to score a single point. This highlights that the structure of a level and the strategy required to solve the "Sokoban" levels can require different approaches to training, even when the mechanics of the game are consistent across levels. 

\subsection{Generalization}

\begin{table}[]
\centering
\begin{tabular}{c|ccc}
                                    & \multicolumn{2}{c}{Evaluation Level}                      \\                
Game                                & 1                             & 3                         \\ \hline
Clusters           &  0.00           $\pm$ 0.00    & 0.7             $\pm$ 0.46                 \\
Cook Me Pasta      &  4.00           $\pm$ 0.00    & 0.00            $\pm$ 0.00                 \\
Bait               & -0.09           $\pm$ 0.29    & 1.78            $\pm$ 0.42                 \\
Sokoban 2          &  0.00           $\pm$ 0.00    & 0.00            $\pm$ 0.00                 \\
Zen Puzzle         & 23.00           $\pm$ 0.00    & 10.9            $\pm$ 5.01                 \\
Labyrinth          &  0.00           $\pm$ 0.00    & \textbf{1.00}   $\pm$ 0.00                 \\
\end{tabular}
\caption{This table shows there results of how well a simple neural network is able to generalize when trained on three levels and evaluated on two unseen levels. The agent can only see an egocentric partially observable view of the environments.}
\label{tab:generalization-results}
\end{table}

The three training levels used in each experiment are 0,1,4 and the evaluation levels are 1 and 3. Table \ref{tab:generalization-results} shows the results of these experiments. These experiments were also trained for 10M steps instead of the 1.28M used in the single environment experiments. In some levels, the agent is able to achieve some rewards, but was only able to solve a single unseen "Labyrinth" level out of the evaluation set. The result on the "Labyrinth" experiment trained per-level also shows that partial observability seems to be less of a challenge for the RL agent than having access to the whole level. The maximum episodic rewards for these environments can be seen in Table \ref{tab:single-env-results}.

\section{Framework Comparison}

\begin{table*}[h]
  \begin{tabular}{llccclllllll}
  \textbf{}                                                 & \multicolumn{1}{l|}{\textbf{}}          & \multicolumn{1}{l}{\textbf{Griddly}} & \multicolumn{1}{l}{\textbf{GVGAI}} & \multicolumn{1}{l}{\textbf{gym-microrts}} & \textbf{MiniGrid} & \textbf{ALE} & \textbf{NLE} & \textbf{DMLab2D} \\ \cline{1-10}
  \multicolumn{1}{c}{\multirow{5}{*}{\textbf{Observation}}} & \multicolumn{1}{l|}{\textbf{Vector}}    & x                                    &                                    & x                                     & x                 & x            & x                & x                \\
  \multicolumn{1}{c}{}                                      & \multicolumn{1}{l|}{\textbf{Block}}     & x                                    &                                    & x                                     & x                 & x            &                  &                  \\
  \multicolumn{1}{c}{}                                      & \multicolumn{1}{l|}{\textbf{Sprites}}   & x                                    & x                                  &                                       &                   &              &                  & x                \\
  \multicolumn{1}{c}{}                                      & \multicolumn{1}{l|}{\textbf{Isometric}} & x                                    &                                    &                                       &                   &              &                  &                  \\
  \multicolumn{1}{c}{}                                      & \multicolumn{1}{l|}{\textbf{ASCII}}     & x                                    & x                                  &                                       &                   &              & x                & x                \\
  \multicolumn{2}{c|}{\textbf{Configurable Partial Observability}}                                    & x                                    &                                    &                                       &                   &              & x                &                  \\
  \multicolumn{2}{l|}{\textbf{GPU Accelerated Rendering}}                                             & x                                    &                                    &                                       &                   &              &                  &                  \\
  \multicolumn{2}{l|}{\textbf{Description Language}}                                                  & x                                    & x                                  &                                       &                   &              &                  & x                \\
  \multicolumn{2}{l|}{\textbf{Procedural Content Generation}}                                         &                                      &                                    &                                       & x                 &              & x                & x                \\
  \multicolumn{2}{l|}{\textbf{Copyable Forward Model}}                                                & x                                    & x                                  & x                                     & x                 & x            &                  &                  \\
  \multirow{3}{*}{\textbf{Player Modes}}                    & \multicolumn{1}{l|}{\textbf{Single}}    & x                                    & x                                  &                                       & x                 & x            & x                & x                \\
                                                            & \multicolumn{1}{l|}{\textbf{Multi}}     & x                                    & x                                  &                                       &                   &              &                  & x                \\
                                                            & \multicolumn{1}{l|}{\textbf{RTS}}       & x                                    &                                    & x                                     &                   &              &                  &    
  \end{tabular}
  \caption{Feature matrix comparing Griddly with other environments. \label{tab:features}}
\end{table*}

In this section we provide two comparisons between several frameworks. The first comparison is a feature matrix showing the differences in features between Griddly and its closest grid-based relatives.  The second comparison is between several games from popular frameworks that have been re-implemented using the GDY language. 

\subsection{Features}

Table~\ref{tab:features} shows how the features offered by Griddly compare with a selection of other environments; since the paper is about Griddly, it is presented to highlight what Griddly offers that other environments do not.  Griddly is most closely related to GVGAI and DMLab2D~\cite{beattie_2020}, but with various extensions to incorporate provide faster rendering and support for multi-agent and grid-based RTS games similar to $\mu$RTS~\cite{ontanon_2013}.

Although ALE is not technically a grid-world, we've
included it in the table for comparison.  NetHack Learning Environment (NLE)
is built around the classic NetHack Rogue-like game, and although just one
grid-based game, it offers great variety due to procedural level generation.  Although not included in the table, ProcGen  
is similar to ALE in scope, but offers endless
variety through procedural level generation.

\begin{table*}

\begin{center}
\resizebox{0.6\linewidth}{!}{%
  \begin{tabular}{ l c c c } 
      Platform          & FPS (Rendered) $\pm$ std. & FPS (Vector)  $\pm$ std.  & Max Memory (MB) \\
      \hline
      \textbf{Griddly}  & 5023 $\pm$ 268             & \textbf{72790} $\pm$ 2474 & 95              \\
      DMLab2D (10x10)   & \textbf{12815} $\pm$ 3863  & 20562 $\pm$ 6658          & \textbf{94}     \\
      \hline
      \textbf{Griddly}  & \textbf{3769} $\pm$ 124    & \textbf{65056} $\pm$ 1736 & 138             \\
      DMLab2D (50x50)   & 984 $\pm$ 116              & 17036 $\pm$ 5341          & \textbf{98}     \\
      \hline
      \textbf{Griddly}  & \textbf{1936} $\pm$ 76     & \textbf{60232} $\pm$ 691  & 371             \\
      DMLab2D (100x100) & 318  $\pm$ 16              & 8577 $\pm$ 2370           & \textbf{146}    \\
      \hline
      \textbf{Griddly}  & \textbf{5012} $\pm$ 244    & \textbf{73134} $\pm$ 839  & \textbf{106}    \\
      GVGAI GYM             & 19   $\pm$ 5               & -                         & 365         \\
      \hline
      \textbf{Griddly}  & \textbf{3799} $\pm$ 170    & \textbf{61101} $\pm$ 4186 & 106             \\
      Minigrid          & 95 $\pm$ 3                 & 1228 $\pm$ 25             & \textbf{49}     \\
      \hline
      \textbf{Griddly}  & \textbf{1160} $\pm$ 157    & \textbf{32130} $\pm$ 419  & \textbf{106}    \\
      gym-microRTS      & 177 $\pm$ 12               & 1906 $\pm$ 272            & 278             \\
      \hline
    \end{tabular}}
\end{center}
\caption{Speed and memory footprint of Griddly compared to similar environments.  All environments are tested using the python OpenAI gym interface except from DMLab2D which has its own equivalent python interface. In each double-row the Griddly entries are for the same or similar game running on each of the platforms.}
\label{tab:compare}
\end{table*}

\subsection{Efficiency}

As the focus of the Griddly Engine is currently to improve the data rate of RL in grid-world environments, a benchmark comparison of the available python gym interfaces for some of most popular grid-based environments is shown in Table \ref{tab:compare}. The benchmark consists of running the original environment and the equivalent Griddly version with a random agent for 1000000 frames and calculating the average frames-per-second (FPS) of the rendered states and maximum memory usage. Rendering the pixels of the environments is the most demanding method of producing game states, so it provides a useful bottleneck to test. Additionally we compare the vectorized versions of the states produced by the game engines if available. The games and maps used for the tests are as follows; GVGAI - Sokoban, MiniGrid - FourRooms, gym-microrts~\cite{huang2020action} - MicrortsMining-v4. We also provide three seperate comparisons to Deepmind 2D lab as it is the most closely related to Griddly. These three comparisons are on three ``Pushbox" game levels with sizes 10x10, 50x50 and 100x100. We also configured the tile size to be the consistent in both Griddly and other platforms.

\section{Future Work}

The baselines in this paper are produced using a simple PPO implementation that was chosen as it was simple to apply to most of the games in the Griddly library. There is an opportunity to test several other more complex algorithms for specific problems such as long-horizon rewards and combinatorial problems.

There are several areas to improve in the Griddly engine itself, for example built-in algorithms for procedural content generation would benefit research into agent generalization. Additionally built-in AI using algorithms such as Monte-Carlo Tree Search could be used for creating baseline adversaries in multi-agent and RTS games.

\section{Conclusion}

In this paper we introduce Griddly, a new, highly efficient framework that allows grid-world games to be easily created using a very flexible description language (GDY) which allows flexibility in desired game mechanics, provides configurable interfaces for various observation, action spaces, level design, and reward functions. This level of flexibility enables a wide range of research possibilities including game playing agents using RL, and level and game design using procedural content generation. The YAML-based GDY language also reduces barriers in language and framework choices as YAML is a widely supported standard.

We provide two sets of RL experiments to make a baseline for future work. The first experiment trains an RL agent separately on different levels of several games with various configurations of observability made possible with Griddly. The second experiment trains a single agent on a few levels of a particular game and evaluates on a few unseen levels. In total a baseline of 156 experiments is given. Both experiment sets show that although the basic RL agents converge, sometimes to the maximum score, the majority of the environments fail to be solved and fail to generalize across new environments. We believe this provides a strong incentive for future experimentation.

\bibliography{references}

\end{document}